\documentclass[runningheads]{llncs}
\usepackage{color,soul}
\usepackage{graphicx}
\usepackage{amsmath}
\usepackage{hyperref}
\usepackage{tcolorbox}

\ifdefined\paperDraft
\def\spcomment#1{{\color{blue}[SP: \textit{#1}]}}
\else
 \def\spcomment#1{}

\fi

\begin{document}
\title{
COVID-19 therapy target discovery with context-aware literature mining}
\titlerunning{COVID-19 therapy target discovery}

\author{Matej Martinc\inst{1,2} \and Bla\v{z} \v{S}krlj\inst{1,2} \and Sergej Pirkmajer \inst{3} \and Nada Lavra\v{c}\inst{1,2,4} 
 \and Bojan Cestnik \inst{5,1} \and Martin Marzidov\v{s}ek \inst{1,2}
\and Senja Pollak \inst{2}}

%
\institute{Jo\v{z}ef Stefan International Postgraduate School, Ljubljana, Slovenia \and
Jo\v{z}ef Stefan Institute, Ljubljana, Slovenia 
\and
Institute of Pathophysiology, Faculty of Medicine, University of Ljubljana, Ljubljana, Slovenia \and
University of Nova Gorica, Vipava, Slovenia \and
Temida d.o.o, Ljubljana, Slovenia
}

\authorrunning{M. Martinc et al.}

\maketitle   

\begin{tcolorbox}

The  final reviewed  publication  was published in Proceedings of the 23rd International Conference on Discovery Science (DS 2020), Thessaloniki, Greece, October 19–21, 2020 and is available online at \url{https://doi.org/10.1007/978-3-030-61527-7_8}.

\end{tcolorbox}

\begin{abstract}
The abundance of literature related to the widespread COVID-19 pandemic is beyond manual inspection of a single expert. Development of systems, capable of automatically processing tens of thousands of scientific publications with the aim to enrich existing empirical evidence with literature-based associations is challenging and relevant. We propose a system for contextualization of empirical expression data by approximating relations between entities, for which representations were learned from one of the largest COVID-19-related literature corpora. In order to exploit a larger scientific context by transfer learning, we propose a novel embedding generation technique that leverages SciBERT language model pretrained on a large multi-domain corpus of scientific publications and fine-tuned for domain adaptation on the CORD-19 dataset. The conducted manual evaluation by the medical expert and the quantitative evaluation based on therapy targets identified in the related work suggest that the proposed method can be successfully employed for COVID-19 therapy target discovery and that it outperforms the baseline FastText method by a large margin.   

\end{abstract}

\section{Introduction}
\label{sec-intro}

Scientific knowledge for a specific domain is in most cases given in an unstructured form, as a set of scientific papers covering a variety of findings, experiments and methodologies related to a specific scientific field or problem. The current speed and quantity of scientific research production makes manual inspection of the literature from a specific field virtually impossible. The recent trend of inter-disciplinary research complicates things even more, as it would require from a researcher to understand all the aspects, from which a specific research problem can be covered in order to ``connect all the dots'' and advance the field by the discovery of the so-called latent scientific knowledge.

To solve this problem, several automated strategies for uncovering this knowledge have been proposed. Somewhat older studies proposed literature-based discovery (LBD)~\cite{bruza2008literature} focusing especially on cross-domain literature mining, which aims at finding interesting bridging terms (b-terms) or bridging links revealing the potentially new connections between separate domain corpora of interest. On the other hand, more recent approaches to latent knowledge discovery from the scientific literature employ word embeddings \cite{mikolov2013efficient}. For example, a study by \cite{tshitoyan2019unsupervised} showed that latent knowledge regarding future discoveries is to a large extent embedded in past publications by retrieving information from the scientific literature with the usage of Word2Vec embeddings \cite{mikolov2013efficient}. 

The latest development in the natural language processing (NLP) is a new type of embeddings called contextual embeddings.  ELMo (Embeddings from Language Models) \cite{peters2018deep} and BERT (Bidirectional Encoder Representations from Transformers) \cite{devlin2018bert} are the most prominent representatives of this type of contextual embeddings, and have been also adapted to scientific literature \cite{beltagy2019scibert}. The main difference between these novel contextual embeddings and older ``static'' embeddings is that in these embeddings a different vector is generated for each context a word appears in, i.e., for each specific word usage in the corpus. These new contextual embeddings solve the problems with word polysemy and other changes in word meaning given different context.  On the other hand, it is not entirely clear how to generate a meaningful general word representation from the word usage embeddings. This means that the usage of contextual embeddings for LBD is not entirely straight forward, since they can not be used in the same way as the traditional static embeddings, and have at least to our knowledge not been used for the task at hand.


In this work, we explore how contextual embeddings can be leveraged for the task of discovering latent scientific knowledge in the very topical scientific literature about the COVID-19 disease. More specifically, we are interested in the discovery of new COVID-19 therapy targets from the targets discovered in the past research. The novelty of this work is two-fold:

\begin{itemize}
\item The paper contributes a new  methodology of generating general word representations from contextual embeddings, proposes an entire workflow for acquisition of novel COVID-19 therapy targets and shows that our method of using contextual embeddings for LBD outperforms the baseline method of using static embeddings by a large margin. 
\item Medically, the paper contributes to identifying new potential COVID-19 therapy targets, motivated by a recent proof-of-concept study that used a state-of-the-art omics approach to identify new possible targets for existing drugs, such as ribavirin \cite{bojkova2020proteomics}.
\end{itemize}


\section{COVID-19 medical background and recent therapy targets}
\label{sec-medicalMotivation}

In late 2019 a novel coronavirus disease (COVID-19), caused by severe acute respiratory syndrome coronavirus 2 (SARS-CoV-2), emerged in China \cite{zhou2020pneumonia,zhu2020novel}. COVID-19 quickly spread and was declared a pandemic by the World Health Organization. 

While new targeted therapies and vaccines against SARS-CoV-2 virus are being actively developed, their potential use in the clinics is not imminent. 
Therefore, until effective pharmacological therapies and/or vaccines are available, medicine needs to resort to other approaches to treat patients with COVID-19 or prevent transmission of SARS-CoV-2. One approach is to identify which among the antiviral drugs that were developed to treat other viral diseases might be effective against SARS-CoV-2. A preliminary report suggests that remdesivir seems to be the most promising candidate among these drugs \cite{beigel2020remdesivir}. Another approach is to identify drugs that are used for other purposes but also exert antiviral effects. The most prominent example among these is hydroxychloroquine, which is used for chronic treatment of rheumatic diseases but also suppresses SARS-CoV-2 in vitro \cite{liu2020hydroxychloroquine}. Identifying a known drug with well-characterized adverse effects would certainly save time and lives before more specific treatments are developed. However, repurposing of existing drugs is also a challenge as highlighted by a recent controversy with hydroxychloroquine \cite{boulware2020randomized,mehra2020cardiovascular} and new candidate drugs and/or therapeutic targets are needed.



\section{Related work}
\label{ref:relatedWork}

The related work is divided into three Sections, namely related work on Literature-based discovery in Section \ref{sec-LBDintro}, related work on text representation learning in Section \ref{sec:representation} and selected overview of recent NLP research on COVID-19 in Section \ref{sec:covid}.

\subsection{Literature-based discovery}
\label{sec-LBDintro}
Literature-based discovery (LBD) aims to generate new knowledge by combining what is already known in the literature. It has been used to (semi-automatically) identify new connections between genes, drugs and diseases, etc. \cite{10.1007/978-3-319-24462-4_8}. Traditionally, LBD has been addressed as finding interesting bridging terms revealing the potentially new connections between separate domain corpora of interest~\cite{bruza2008literature}. Swanson \cite{swanson1990medical} developed one of the early LBD approaches, the so-called ABC model, to detecting interesting b-terms to uncover the possible cross-domain relations among previously unrelated concepts. 

On the other hand, a more recent state-of-the art tool LION LBD \cite{10.1093/bioinformatics/bty845} enables researchers to navigate published information and supports hypothesis generation and testing. The system is built with a particular focus on the molecular biology of cancer. LBD has led to discovery of potential treatments in other domains, including multiple sclerosis \cite{KOSTOFF2008239}, and has been applied successfully in drug development and repurpusing \cite{Deftereos2011}. 
Recent LBD approaches benefit from word embeddings. One is the study by \cite{tshitoyan2019unsupervised} already mentioned in Section \ref{sec-intro} and the other is the work by \cite{neuralLBD}, who proposed graph-based, neural network methods to perform open and closed LBD and demonstrated improved performance on existing tasks.

\subsection{Text representation and embeddings}
\label{sec:representation}

Recently, the embedding approach became a prevalent way to build representations for many different types of entities, e.g., texts, graphs, electronic health records, images, relations, recommendations, etc. Text embeddings use large corpora of documents to extract vector representations for words, sentences, and documents. 
The first neural word embeddings like Word2vec \cite{mikolov2013efficient} produced one vector for each word, irrespective of its polysemy. These so-called static embeddings have been further developed and the most popular static embeddings currently in use besides Word2Vec are GloVe (Global vectors for word representation) \cite{pennington2014glove} and FastText \cite{Bojanowski_2017}.
Recent developments like ELMo \cite{peters2018deep} and BERT \cite{devlin2018bert} take a context of a sentence into account and produce different word vectors for different contexts of each word. Another novelty of these approaches is the employment of the transfer learning technique, which has recently become a well established procedure in the field of NLP. This procedure relies on a language model pretraining on very large unlabeled textual resources and after that transfer of the knowledge obtained by the language model onto a specific downstream task by further fine-tuning the model.

\subsection{Text mining and NLP research related to COVID-19}
\label{sec:covid}

With regard to biomedical research on COVID-19, time is a central factor as scientists try to design treatments and vaccines amid the pandemic caused by the SARS-CoV-2 virus, therefore leveraging LBD and its potential to reduce scientific discovery time could prove crucial. 

Many search platforms emerged for retrieving COVID-19 related papers. For example, Neural Covidex\footnote{\url{https://covidex.ai/}} is based on neural ranking architecture and provides information access capabilities to the COVID-19 Open Research Dataset (CORD-19) (see Section \ref{sec:literature-considered}). SciSight \cite{hope2020scisight} in contrast to standard targeted search facilitates finding connections between biomedical concepts that are not obvious from reading individual papers. It displays a network of top related terms mined from the corpus, based on the co-appearance in the same sentence. 

Studies that can generate new knowledge about COVID-19 by applying embeddings are still scarce but do exist. For example, a recent study has projected Covid-related medical texts in a 3D human atlas space that helps to navigate the literature \cite{grujicic2020self}. The objective was to learn semantically aware groundings of sentences with five different BERT models \cite{devlin2018bert}.


\section{Background knowledge and resources}
\label{sec:resources}

We describe the CORD-19 corpus (Section \ref{sec:literature-considered}) and embeddings technology (Section \ref{sec:tmbackground}) used in this study.

\subsection{CORD19 database}
\label{sec:literature-considered}
The scientific literature considered in this work has been recently introduced as the CORD-19 corpus\footnote{\url{https://www.kaggle.com/allen-institute-for-ai/CORD-19-research-challenge}}. CORD-19 is a resource of over 135{,}000 scholarly articles, including over 68{,}000 with full text, about COVID-19, SARS-CoV-2, and related coronaviruses. This freely available data set is provided to the global research community to apply recent advances in NLP and other AI techniques to generate new insights in support of the ongoing fight against this infectious disease. 

We use the corpus version 12, published on May 1st 2020, from which we extract only full text scholarly articles converted into xml from a pdf format. This results in altogether 48,410 papers, which are summarized in Table \ref{tbl:dataset}.

\begin{table}[!h]
\centering
\caption{CORD-19 dataset statistics.}
\label{tbl:dataset}
\resizebox{0.99\textwidth}{!}{
\begin{tabular}{|l|c|c|}
    \hline
    Origin & Number of papers & Number of tokens \\\hline 
    Commercial use subset & 9,918 & 46,206,453 \\
    Non-commercial use subset & 2,584 & 10,732,608 \\
    PMC custom license subset & 32,450 & 156,247,363 \\
    bioRxiv (not peers reviewed) & 2,670 & 8,968,183 \\
    medRxiv subset (not peer reviewed) & 788 & 3,285,558 \\\hline
    All & 48,410 & 225,440,165\\
    \hline
\end{tabular}
}
\end{table}

\subsection{Considered embeddings}
\label{sec:tmbackground}

We use FastText \cite{Bojanowski_2017} embeddings as a baseline in this study. The main advantage of FastText embeddings is its word representation as a sum of n-grams, which allows the model to, in addition to leveraging semantic relations, also leverage morphological information.

One of the most oftenly used models for the generation of contextual embeddings is the BERT model \cite{devlin2018bert} that was originally pretrained on the Google Books Corpus (800 million tokens) and Wikipedia (2,500 million tokens). This pretraining is however not entirely appropriate for the text mining tasks on the scientific literature due to specificities of the scientific language and vocabulary. For this reason, in this research we opted for SciBERT \cite{beltagy2019scibert}, a version of BERT pretrained on a large multi-domain corpus of scientific publications, a random sample of 1.14M papers from Semantic Scholar. SciBERT model has 12 encoder layers with the attention mechanism and a hidden layer size of 768.

\section{Methodology}
\label{sec:Methodology}
In this section, we present the methodology of the proposed approach by explaining how we obtain word representations, how we acquire therapy target candidates and how we evaluate the approach.

\subsection{Word representations}
\label{sec:wordRep}

\begin{figure*}[!h]
\begin{center}
\includegraphics[width = 0.54\linewidth]{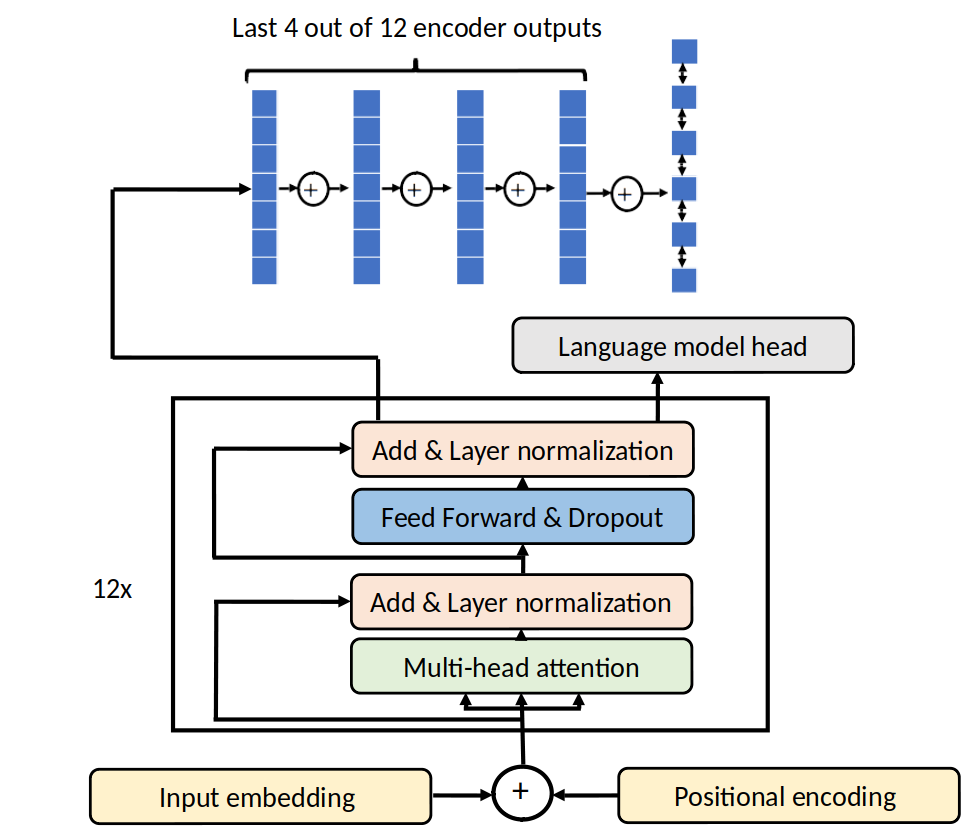}
\caption{Extraction of word usage embeddings from BERT. Note that only the last 4 out of 12 BERT encoder layers are used for the embedding generation. This was done in accordance with the previous studies that suggested that the last four layers carry the bulk of the semantic information obtained by the model \cite{martinc2019leveraging}.}
\label{fig.bert}
\end{center}
\end{figure*}

First, we fine-tune SciBERT as a masked language model for domain adaptation on the lowercased CORD-19 dataset. Next, we generate word representations for each word in the vocabulary. Figure \ref{fig.bert} visualizes the process described below. The documents from the corpus are split into sequences of byte-pair encoded tokens \cite{kudo2018sentencepiece} of a maximum length of 256 tokens and fed into the fine-tuned SciBERT model. For each of these sequences of length $n$, we create a sequence embedding by summing the last four encoder output layers. The resulting sequence embedding of size $n$ times \textit{embeddings size} represents a concatenation of contextual embeddings for the $n$ tokens in the input sequence. By chopping it into $n$ pieces, we acquire a representation, i.e. a contextual token embedding, for each word used in the corpus. Note that these representations vary according to the context in which the token appears, meaning that the same word has a different representation in each specific context (sequence). 

Finally, the resulting embeddings are aggregated on the token level (i.e. for every token in the corpus vocabulary, we create a list of all their contextual embeddings) and are averaged, in order to get one representation for each token in the vocabulary. We enforce a constraint that a list of contextual embeddings for a specific token should contain at least five elements, otherwise the specific token is discarded. This is done in order to remove tokens that do no appear in the corpus enough times for the model to learn a meaningful representation (e.g., mostly tokens that contain typos or very rare technical terms).  
Since the byte-pair input encoding scheme \cite{kudo2018sentencepiece} employed by the SciBERT model does not necessarily generate tokens that correspond to words but rather generate tokens that correspond to parts of words, we also propose the following \textit{on the fly} reconstruction mechanism that allows us to get word representations from byte pair tokens. If a word is split into more than one byte pair token, we take an embedding for each byte pair token constituting a word and build a word embedding by averaging these byte pair tokens. The resulting average is used as a context specific word representation.

The final result are static embeddings for each word in the vocabulary, capable of leveraging a broader semantic knowledge due to the SciBERT being pretrained on a large corpus of scientific articles.  
As a baseline, we also train a FastText skip-gram model with an embedding dimension of 100 (which is the default) on the lowercased CORD-19 dataset. Once again we enforce the constraint that a word should appear in the corpus at least five times.

\subsection{Synonym resolution}
\label{sec:synonyms}

Once embeddings are generated, we conduct synonym resolution with the help of a list of 19,302 gene names and their most common synonyms \cite{povey2001hugo}. The embedddings belonging to the synonyms of the same gene are averaged in order to combine contextual information of different identifiers referring to the same gene and in order to avoid possible mismatches due to different naming. 

\subsection{Candidate acquisition}
\label{sec:acquisition}

The main idea of our approach is to leverage semantic similarity in order to derive new scientific knowledge from an already existing one. For this to work, some initial seed concepts need to be acquired and used as a starting point. We explore two possibilities for this:

\begin{itemize}
    \item \textbf{Seed concepts recommended by the expert}: The experts with a medical background were asked to recommend genes and/or proteins with a known and confirmed link to COVID-19. The final consensus was to focus on angiotensin-converting enzyme 2 (ACE2) and transmembrane protease serine 2 (TMPRSS2). ACE2, a receptor for the spike S protein, is important because SARS-CoV-2 uses it to enter the host cell \cite{hoffmann2020sars}. TMPRSS2 promotes SARS-CoV-2 entry into the cell by priming the spike S protein \cite{hoffmann2020sars}. Blockage of binding of SARS-CoV-2 to ACE2 or inhibition of TMPRSS2 are therefore two possible approaches to treat COVID-19. 
    \item \textbf{Seed concepts found in the literature}: Due to the abundance of recent research on COVID-19 it is also possible to find seed concepts in the related research. We opted for a study by \cite{bojkova2020proteomics} in which a set of COVID-19 therapy targets were identified. The considered list of altogether 2802 potential targets\footnote{Note that the original list contains 2715 targets (see Supplementary Table 1 in \cite{bojkova2020proteomics}). Some of them are however represented as a set of similar genes/proteins belonging to the same family. On the other hand, we treat each individual gene/protein as a separate target, which results in a set of 2802 targets.} is the result of a large-scale screening for active proteins, and offers a starting set of candidates obtained empirically. The list is ranked according to the increase or decrease of production of a specific protein at a specific time point. We explore what is the optimal number of seed candidates by exponentially enlarging the size of the seed candidate set. Sampling from the list is conducted according to the ranking of the protein candidates, i.e., we sample 2, 4, 8, 16, 32 and 64 best ranked seed candidates according to the increase in their production 24 hours after the infection (column Ratio 24h in Supplementary Table 1 in the study by \cite{bojkova2020proteomics}).
\end{itemize}

Once seed concepts are acquired, we calculate their embeddings and look for semantically similar concepts by finding the concepts that are the closest to seed concepts according to the cosine distance between the embeddings\footnote{Note that these concepts obtained according to semantic similarity are not necessarily proteins/genes but rather any word in the embedding vocabulary.}. More specifically, we find a set of 2802 closest candidate concepts for each gene/protein in each seed candidate set, and the acquired candidates are ranked according to the cosine similarity. Finally, we calculate the average ranking for each candidate (i.e. by averaging ranks for each seed concept in the set) and therefore obtain $\textrm{NumOfCandidatesInSet}*2802$ closest candidates for each of the seed concept sets with possible duplicates originating from different seed concepts. 

Since the initial experiments showed that many of the most similar concepts are in fact variations of the same base concept (e.g., the closest neighbours to ACE2 being ACE, ACE2M, ACE2S...) and since we are interested in maximizing the variety of the acquired candidates, we conduct an additional filtering according to the normalized Levenshtein distance defined as:

\begin{equation*}
  \textsc{normLD} = 1 - \frac{LD}{ \max(\textrm{len}(w_1),\textrm{len}(w_2))} ,
\end{equation*}

\noindent where $\textsc{normLD}$ stands for normalized Levenshtein distance, $LD$ for Levenshtein distance, $w_1$ is either a seed concept or a concept already in the list of acquired neighbours and $w_2$ is the new candidate neighbour. Concepts for which normalized Levenshtein difference is bigger than 0.7 are discarded\footnote{The normalized Levenshtein difference threshold of 0.7 was chosen empirically.}. The filtering is conducted in order from the top of the list (neighbours with the best average rank) to the bottom.

At the end of the candidate acquisition process, we cut the ranked list of neighbours at 2802 target candidates for each of the distinct seed concept sets used in the evaluation.

\subsection{Evaluation}

The methods for discovering new therapy targets are evaluated in two evaluation settings, quantitative and qualitative. 

\subsubsection{Quantitative Evaluation}
\label{sec:quantitativeEvalMethodology}

We evaluate if therapy target candidates acquired in the previous step have been confirmed as targets in the study by \cite{bojkova2020proteomics}, i.e. how many of them appear in the list of 2802 candidates they identified\footnote{Note that the study by \cite{bojkova2020proteomics} is not included in the CORD-19 corpus used for training the embeddings, since it was published on May 14th 2020 and we use the CORD-19 version published on May 1st 2020.}. Note that in this setting we only evaluate the proposed method on the previously existing knowledge, therefore in the quantitative evaluation we can not asses, if the method has managed to discover some potentially useful and previously undiscovered knowledge. 

We are interested in precision at rank $k$. This means that only the candidates ranked equal to or higher than $k$ are considered and the rest are disregarded. Precision is the ratio of the number of relevant candidates divided by the number of candidates returned by the system, or more formally:

\[\textrm{precision} = \frac{|\textrm{relevant}~~ \textrm{candidates}@k|}{|\textrm{returned}~~ \textrm{candidates}|}\]

Recall@$k$ is the ratio of the number of relevant candidates ranked equal to or higher than $k$ by the system divided by the number of correct ground truth candidates:
    
\[\textrm{recall} = \frac{|\textrm{relevant}~~ \textrm{candidates}@k|}{|\textrm{correct}~~ \textrm{candidates}|}\]
 
We measure precision and recall at k=100 and k=2802 in order to investigate how different number of retrieved candidates for each seed concept set affects the precision and recall of the methods. More specifically, we are trying to confirm or deny a hypothesis that larger k values degrade the overall precision of the method. 

The relevance of the candidate is determined according to two matching criteria. First one is the \textbf{exact} match, where the candidate is deemed relevant if it appears in the list of identified targets in the study by \cite{bojkova2020proteomics}. The second is the \textbf{fuzzy} match, where we check if the targets belong to the same ``family'' as a specific confirmed target. This strategy was proposed by the medical experts and checks whether the prefix of the specific gene (characters in the gene name that appear before the first digit in the name) matches a prefix of a specific gene name in the list. We enforce an additional constraint that the matching prefixes need to be at least three characters long for a successful match in order to minimize the false positive rate.

\subsubsection{Qualitative Evaluation}

We generated two distinct therapy target candidate lists using the proposed SciBERT based embedding method. First one contained 100 closest neighbours to the protein ACE2 according to the cosine distance between embeddings, and the second one contained 100 closest neighbours to the protein TMPRSS2. Both lists were given to the medical expert who inspected the list for possible previously undiscovered candidates.

\section{Results}
\label{sec:results}

Here we present the results of the quantitative and qualitative evaluation.

\subsection{Results of the quantitative evaluation}

The results of the quantitative evaluation are presented in Table \ref{tbl:quantitativeEval}. In column ACE2 + TMPRSS2 we present results when these two proteins are used as seed concepts, and in column UBA2 + NCKAP1 we present results when these two proteins, which were chosen according to the largest value of the Ratio 24h criterion (see Section \ref{sec:acquisition}) are used as seed concepts. Left part of the Table presents results for the proposed approach based on SciBERT and the right part of the Table presents results for the baseline FastText approach in terms of precision and recall at two distinct $k$ values (100 and 2802). EXACT indicates that exact matching is used and FUZZY indicates fuzzy matching (see Section \ref{sec:quantitativeEvalMethodology}). 

\begin{table}[!b]
\centering
\caption{Results (precision@k and recall@k) of the quantitative evaluation for two seeds by the expert and two seeds from the literature. Best result in each row is bolded.}
\label{tbl:quantitativeEval}
\resizebox{0.99\textwidth}{!}{
\begin{tabular}{|l|c|c|c|c|}
    \hline
    & \multicolumn{2}{c|}{\textbf{SciBERT}} & \multicolumn{2}{c|}{\textbf{FastText}} \\
    & ACE2 + TMPRSS2 & UBA2 + NCKAP1 & ACE2 + TMPRSS2 & UBA2 + NCKAP1\\
    \hline
    EXACT P@100 & 0.110 & \textbf{0.220} & 0.040 & 0.170\\
    EXACT R@100 & 0.004 & \textbf{0.008} & 0.001 & 0.006\\
    EXACT P@2802 & 0.097 & \textbf{0.118} & 0.025 & 0.076\\
    EXACT R@2802 & 0.097 & \textbf{0.118} & 0.025 & 0.076\\
    \hline
    FUZZY P@100 & 0.290 & \textbf{0.490} & 0.070 & 0.380\\
    FUZZY R@100 & 0.010 & \textbf{0.017} & 0.002 & 0.014\\
    FUZZY P@2802 & 0.222 & \textbf{0.252} & 0.092 & 0.183\\
    FUZZY R@2802 & 0.222 & \textbf{0.252} & 0.092 & 0.183\\
    \hline
\end{tabular}
}
\end{table}

SciBERT based method outperforms the FastText baseline by a large margin in both seed therapy target acquisition scenarios and according to all the criteria. Using UBA2 + NCKAP1 works better than using ACE2 + TMPRSS2, achieving the best fuzzy precision@100  of 0.490 and the best exact precision@100 of 0.220. FastText baseline also works fairly well in this scenario, achieving fuzzy precision@100 of 0.380 and the best exact precision@100 of 0.170. When more (2802) candidates are obtained, the recall increases for both methods but at an expense of a significant drop in precision for both methods and for almost all configurations. The only exception is the increase in fuzzy precision by about 2 percentage points when FastText method and ACE2 + TMPRSS2 seed concepts are used. The most likely reason for the drop is that at larger k values some of the target candidates acquired by the method might be semantically too dissimilar to the seed targets, since more candidates per each seed therapy target need to be acquired in order to get the required amount of semantic neighbours (e.g., for k=2802, we get about 1401 semantic neighbours for each of the seed genes).

\begin{figure*}[!t]
\begin{center}
\includegraphics[width = 1.0\linewidth]{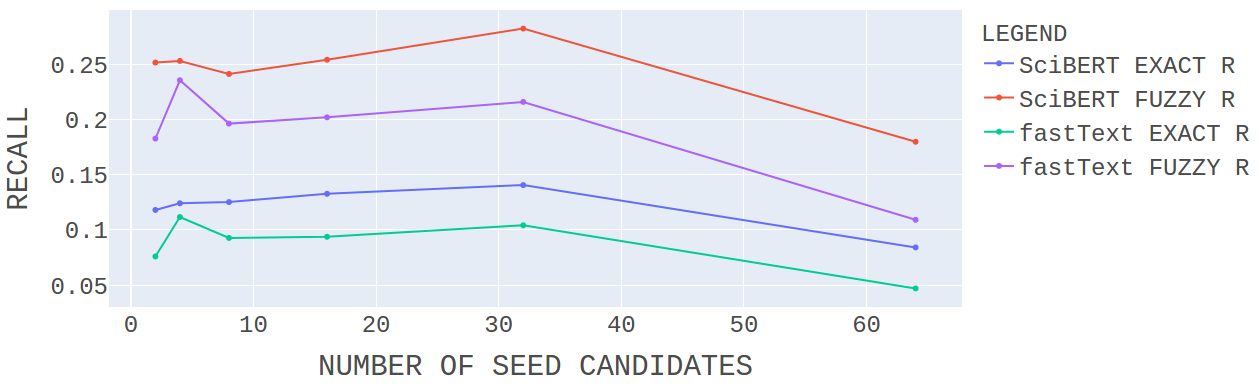}
\caption{Relation between recall and the number of seed candidates.}
\label{fig.seedCand}
\end{center}
\end{figure*}

This raises the question of how many seed terms should be supplied to the system for the best performance when a large number of target candidates is required as output. Figure \ref{fig.seedCand} shows the relation between the achieved recall@2802 (exact and fuzzy) of both methods when we increase the number of seed candidates (see Section \ref{sec:acquisition} for details about our sampling procedure). For SciBERT based method, the best fuzzy and exact recalls are achieved when 32 seed candidates are used (28.2\% and 14.1\% respectively). On the other hand, the FastText based method shows a spike in performance when 4 seed concepts are used. This indicates that for some reason the two seed candidates ranked third and fourth (ENO1 and ATP5O, respectively) according to the Ratio 24 criterion have a very positive effect on the FastText model but not SciBERT. While we do not have a clear explanation for this phenomenon, it is hypothesized that it might be connected with morphological similarity between these two genes and other genes in the list of candidates proposed by \cite{bojkova2020proteomics}, since FastText can also leverage morphological similarity. Spikes asides, the general trend for both methods and both recalls is quite similar. There is a gradual increase in performance for up to 32 seed candidates and after that the performance decreases. 

\subsection{Results of the qualitative evaluation}

Nine genes/proteins were the same in the ACE2 and TMPRSS2 lists, indicating they might be important for pathogenesis of COVID-19. The role of these genes/proteins in pathogenesis of COVID-19 has not been established, but indirect evidence supports this notion at least for some of them. Indeed, most of these genes/proteins have been previously linked to viral diseases, including those caused by SARS-CoV (a virus, which causes SARS, and is related to SARS-CoV-2), and other coronaviruses (Table \ref{tbl:genes}). Furthermore, METAP2 and DPP7, which we identified as potentially relevant for COVID-19, were altered in cells infected with SARS-CoV-2, although the difference for DPP7 did not reach the level of statistical significance \cite{bojkova2020proteomics}.

Interestingly, three proteins in Table \ref{tbl:genes} (PTGS2, CRTH2, and PLA2R1) are linked to infection with coronaviruses as well as metabolism of phospholipids and/or prostaglandin synthesis and action. Furthermore, both the ACE2 and TMPRSS2 lists contain genes/proteins, such as PLA2 (phospholipase A2, PLA2G2D 
(Group IID secretory phospholipase A2), and SPLA2 (secretory PLA2), which do not match directly, but are involved in the same or related cellular processes. Notably, increased expression of \textit{Pla2g2d} in older mice was shown to be linked with increased mortality due to SARS-CoV infection \cite{vijay2015critical}. In addition, a recent proteomic analysis has demonstrated that protein abundance of PLAA (phospholipase A2-activating protein), PLA2G4A (cytosolic phospholipase A2), and PLA2G2 (Group IIA phospholipase A2) is altered in cultured cells infected with SARS-CoV-2 \cite{bojkova2020proteomics}, which gives further credence to the idea that phospholipid metabolism is important under these conditions. In summary, taken together with published experimental data, our analysis suggests that phospholipases and/or prostaglandins might represent a target for treatment of COVID-19.

\begin{table}[!t]
\centering
\caption{Genes/proteins (in alphabetical order) which are common to the TMPRSS2 and ACE2 list and their (putative) relevance to COVID-19.}
\label{tbl:genes}
\resizebox{0.99\textwidth}{!}{
\begin{tabular}{|c|c|c|}
    \hline
    \textbf{Gene} & \textbf{Protein} & \textbf{Relevance to COVID-19} \\
    \hline
    \textit{ATP2B2 (PMCA2)} & Plasma membrane $Ca^{2+}$-transporting ATPase & ?\\
    \hline
    \textit{CRTH2 (PTGDR2)} & Prostaglandin D2 (PGD2) receptor &  \begin{tabular}{@{}c@{}}PGD2 is important for survival of mice infected \\ with neurotropic coronavirus. Increased production of PGD2 \\is linked to increased mortality in aged mice. PGD2blockade \\ improves survival in mice infected with SARS-CoV \cite{vijay2017virus,zhao2011age}. \end{tabular} \\
    \hline
    \textit{DPP7 (DPP2)} & Dipeptidyl peptidase 2 & \begin{tabular}{@{}c@{}}DPP7 is associated with the magnitude of \\ the antibody response to influenza vaccination \cite{hipc2017multicohort}. \end{tabular}\\ 
    \hline
   \textit{MECP2} & Methyl-CpG-binding protein 2 & \begin{tabular}{@{}c@{}}MECP2 duplication in humans is associated with IgA/IgG2 \\ antibody deficiency and severe infections. Mice overexpressing \\ MECP2 are hypersensitive to influenza A virus \cite{bauer2015infectious,cronk2017influenza}.\end{tabular}\\
    \hline
    \textit{METAP2 (P67EIF2)} & \begin{tabular}{@{}c@{}}Methionine aminopeptidase 2 (Initiation \\ factor 2-associated 67 kDa glycoprotein) \end{tabular} & \begin{tabular}{@{}c@{}}Plays a role in regulation of protein\\ synthesis during vaccinia virus infection \cite{bose1997viral}.\end{tabular}\\
    \hline
    \textit{PLA2R1} & Secretory phospholipase A2 (PLA2) receptor & \begin{tabular}{@{}c@{}}Restricted activity of PLA2 is associated with improved \\ survival in mice infected with HCoV-OC43. 
Inhibition \\ of cytosolic PLA2 suppresses replication of HCoV-229E \cite{do2008neuroprotective,Muller2018}.\end{tabular}\\
    \hline 
    \textit{PTGS2 (COX2)} & Prostaglandin G/H synthase 2 (cyclooxygenase-2) & SARS-CoV induces cyclooxygenase 2 \cite{liu2006amino}. \\
    \hline
    \textit{SOX2} & Transcription factor SOX-2 & \begin{tabular}{@{}c@{}}SOX2+ cells are important for regeneration of airway\\ epithelium after severe influenza infection in mice \cite{ray2016rare}.\end{tabular}\\
    \hline
    \textit{SSTR2 (SST2)} & Somatostatin receptor type 2 & ?\\
    \hline
\end{tabular}
}
\end{table}

\section{Conclusions and further work}
\label{sec:conclusion}

In this paper we presented a method for discovering new COVID-19 therapy targets by leveraging contextual embeddings, which outperforms the method based on FastText embeddings. We explored the best tactics for acquiring seed targets from the related work if expert knowledge is not available. The results of the manual qualitative evaluation by the expert indicate that at least two groups of novel therapy target candidates have been discovered.

The proposed method outperforms the baseline FastText method by a large margin, which can be explained by the fact that SciBERT is also leveraging knowledge gained during the pretraining on the large corpus of scientific literature, which enables the model to generate vector representations that reflect this wider semantic context. The drawback is however the difference in the amount of computational resources required by the two methods. We also acknowledge that the proposed method, which constructs static embeddings from the SciBERT contextual embeddings is not the only possibility for construction of meaningful semantic representations. Other possibilities and models (e.g., BioBERT \cite{lee2020biobert}) will be explored in the future work.
The quantitative evaluation indicates that the precision and recall of the method are still relatively low in most cases. This can on one side indicate that COVID-19 topic is not researched enough to confirm relations between COVID-19 and some candidates found by the proposed method. Another indication of this is the qualitative study, which confirmed that some of the proposed candidates found by the system have research potential but have not yet been explicitly confirmed as being related to COVID-19 in the existing literature. 

On the other hand, low precision most likely also indicates that there is still a large amount of proposed candidates, which play no role in the advancement and prevention of the COVID-19 disease. Some of these false positives can be attributed to inadequate synonym resolution since the list used for that task (see Section \ref{sec:synonyms}) most likely covers only a small percentage of genes and their synonyms found in the CORD-19 corpus. Other mistakes can be contributed to the byte pair encoding scheme SciBERT employs. Since the model generates embeddings for subword tokens instead for an entire words (see how we deal with this problem in Section \ref{sec:wordRep}), some words with similar roots or affixes can perhaps appear closer in the semantic space as they should according to their semantic relatedness because of the morphological resemblance. We will address this issues in the future work.

\subsubsection*{Acknowledgements}
\footnotesize{ This work was funded by the Slovenian Research Agency (ARRS) through core research programme \emph{Knowledge Technologies} (P2-0103), research project \emph{Semantic Data Mining for Linked Open Data} (financed under the ERC Complementary Scheme, N2-0078), and a young researcher grant (BŠ). The work was also supported by EU Horizon 2020 research and  innovation programme under grant agreement No 825153, project EMBEDDIA (Cross-Lingual Embeddings for
Less-Represented Languages in European News Media). The publication reflects only the authors' views and the EC is not responsible for any use that may be made of the information it contains.}
\bibliographystyle{splncs04}

\bibliography{lbd,textmining,sergej,LBD-covid}

\begin{thebibliography}{10}
\providecommand{\url}[1]{\texttt{#1}}
\providecommand{\urlprefix}{URL }
\providecommand{\doi}[1]{https://doi.org/#1}

\bibitem{bauer2015infectious}
Bauer, M., K{\"o}lsch, U., Kr{\"u}ger, R., Unterwalder, N., Hameister, K.,
  Kaiser, F.M., Vignoli, A., Rossi, R., Botella, M.P., Budisteanu, M., et~al.:
  Infectious and immunologic phenotype of mecp2 duplication syndrome. Journal
  of Clinical Immunology  \textbf{35}(2),  168--181 (2015)

\bibitem{beigel2020remdesivir}
Beigel, J.H., Tomashek, K.M., Dodd, L.E., Mehta, A.K., Zingman, B.S., Kalil,
  A.C., Hohmann, E., Chu, H.Y., Luetkemeyer, A., Kline, S., et~al.: Remdesivir
  for the treatment of covid-19—preliminary report. New England Journal of
  Medicine  (2020)

\bibitem{beltagy2019scibert}
Beltagy, I., Cohan, A., Lo, K.: Scibert: Pretrained contextualized embeddings
  for scientific text. arXiv preprint arXiv:1903.10676  (2019)

\bibitem{Bojanowski_2017}
Bojanowski, P., Grave, E., Joulin, A., Mikolov, T.: Enriching word vectors with
  subword information. Transactions of the Association for Computational
  Linguistics  \textbf{5},  135–146 (Dec 2017)

\bibitem{bojkova2020proteomics}
Bojkova, D., Klann, K., Koch, B., Widera, M., Krause, D., Ciesek, S., Cinatl,
  J., M{\"u}nch, C.: Proteomics of sars-cov-2-infected host cells reveals
  therapy targets. Nature pp.~1--8 (May 2020). \doi{10.1038/s41586-020-2332-7}

\bibitem{bose1997viral}
Bose, A., Saha, D., Gupta, N.K.: Viral infection: I. {R}egulation of protein
  synthesis during vaccinia viral infection of animal cells. Archives of
  Biochemistry and Biophysics  \textbf{342}(2),  362--372 (1997)

\bibitem{boulware2020randomized}
Boulware, D.R., Pullen, M.F., Bangdiwala, A.S., Pastick, K.A., Lofgren, S.M.,
  Okafor, E.C., Skipper, C.P., Nascene, A.A., Nicol, M.R., Abassi, M., et~al.:
  A randomized trial of hydroxychloroquine as postexposure prophylaxis for
  covid-19. New England Journal of Medicine  (2020)

\bibitem{bruza2008literature}
Bruza, P., Weeber, M.: Literature-based Discovery. Springer Science \& Business
  Media (2008)

\bibitem{neuralLBD}
Crichton, G., Baker, S., Guo, Y., Korhonen, A.: Neural networks for open and
  closed literature-based discovery. PLOS ONE  \textbf{15}(5),  1--16 (05 2020)

\bibitem{cronk2017influenza}
Cronk, J.C., Herz, J., Kim, T.S., Louveau, A., Moser, E.K., Sharma, A.K.,
  Smirnov, I., Tung, K.S., Braciale, T.J., Kipnis, J.: Influenza a induces
  dysfunctional immunity and death in mecp2-overexpressing mice. JCI Insight
  \textbf{2}(2) (2017)

\bibitem{Deftereos2011}
Deftereos, S.N., Andronis, C., Friedla, E.J., Persidis, A., Persidis, A.: Drug
  repurposing and adverse event prediction using high-throughput literature
  analysis. Wiley Interdisc. Reviews: Systems Biology and Medicine
  \textbf{3}(3),  323--334 (2011)

\bibitem{devlin2018bert}
Devlin, J., Chang, M.W., Lee, K., Toutanova, K.: Bert: Pre-training of deep
  bidirectional transformers for language understanding. arXiv preprint:
  1810.04805  (2018)

\bibitem{do2008neuroprotective}
Do~Carmo, S., Jacomy, H., Talbot, P.J., Rassart, E.: Neuroprotective effect of
  apolipoprotein d against human coronavirus oc43-induced encephalitis in mice.
  Journal of Neuroscience  \textbf{28}(41),  10330--10338 (2008)

\bibitem{grujicic2020self}
Grujicic, D., Radevski, G., Tuytelaars, T., Blaschko, M.B.: Self-supervised
  context-aware covid-19 document exploration through atlas grounding  (2020)

\bibitem{hipc2017multicohort}
{HIPC-I Consortium}, et~al.: Multicohort analysis reveals baseline
  transcriptional predictors of influenza vaccination responses. Science
  immunology  \textbf{2}(14) (2017)

\bibitem{hoffmann2020sars}
Hoffmann, M., Kleine-Weber, H., Schroeder, S., Kr{\"u}ger, N., Herrler, T.,
  Erichsen, S., Schiergens, T.S., Herrler, G., Wu, N.H., Nitsche, A., et~al.:
  Sars-cov-2 cell entry depends on ace2 and tmprss2 and is blocked by a
  clinically proven protease inhibitor. Cell  (2020)

\bibitem{hope2020scisight}
Hope, T., Portenoy, J., Vasan, K., Borchardt, J., Horvitz, E., Weld, D.S.,
  Hearst, M.A., West, J.: Scisight: Combining faceted navigation and research
  group detection for covid-19 exploratory scientific search. arXiv preprint:
  2005.12668  (2020)

\bibitem{10.1007/978-3-319-24462-4_8}
Korhonen, A., Guo, Y., Baker, S., Yetisgen-Yildiz, M., Stenius, U., Narita, M.,
  Li{\`o}, P.: Improving literature-based discovery with advanced text mining.
  In: DI~Serio, C., Li{\`o}, P., Nonis, A., Tagliaferri, R. (eds.)
  Computational Intelligence Methods for Bioinformatics and Biostatistics. pp.
  89--98. Springer, Cham (2015)

\bibitem{KOSTOFF2008239}
Kostoff, R.N., Briggs, M.B., Lyons, T.J.: Literature-related discovery ({LRD}):
  Potential treatments for multiple sclerosis. Technological Forecasting and
  Social Change  \textbf{75}(2),  239--255 (2008)

\bibitem{kudo2018sentencepiece}
Kudo, T., Richardson, J.: Sentencepiece: A simple and language independent
  subword tokenizer and detokenizer for neural text processing. arXiv
  preprint:1808.06226  (2018)

\bibitem{lee2020biobert}
Lee, J., Yoon, W., Kim, S., Kim, D., Kim, S., So, C.H., Kang, J.: {BioBERT}: a
  pre-trained biomedical language representation model for biomedical text
  mining. Bioinformatics  \textbf{36}(4),  1234--1240 (2020)

\bibitem{liu2020hydroxychloroquine}
Liu, J., Cao, R., Xu, M., Wang, X., Zhang, H., Hu, H., Li, Y., Hu, Z., Zhong,
  W., Wang, M.: Hydroxychloroquine, a less toxic derivative of chloroquine, is
  effective in inhibiting sars-cov-2 infection in vitro. Cell Discovery
  \textbf{6}(1), ~1--4 (2020)

\bibitem{liu2006amino}
Liu, M., Gu, C., Wu, J., Zhu, Y.: Amino acids 1 to 422 of the spike protein of
  sars associated coronavirus are required for induction of cyclooxygenase-2.
  Virus genes  \textbf{33}(3),  309--317 (2006)

\bibitem{martinc2019leveraging}
Martinc, M., Novak, P.K., Pollak, S.: Leveraging contextual embeddings for
  detecting diachronic semantic shift. arXiv preprint arXiv:1912.01072  (2019)

\bibitem{mehra2020cardiovascular}
Mehra, M.R., Desai, S.S., Kuy, S., Henry, T.D., Patel, A.N.: Retraction:
  Cardiovascular disease, drug therapy, and mortality in covid-19. New England
  Journal of Medicine  (2020)

\bibitem{mikolov2013efficient}
Mikolov, T., Chen, K., Corrado, G., Dean, J.: Efficient estimation of word
  representations in vector space. arXiv preprint arXiv:1301.3781  (2013)

\bibitem{Muller2018}
M{\"u}ller, C., Hardt, M., Schwudke, D., Neuman, B.W., Pleschka, S., Ziebuhr,
  J.: Inhibition of cytosolic phospholipase a2$\alpha$ impairs an early step of
  coronavirus replication in cell culture. Journal of virology  \textbf{92}(4),
   e01463--17 (2018)

\bibitem{pennington2014glove}
Pennington, J., Socher, R., Manning, C.D.: Glove: Global vectors for word
  representation. In: Proc. of the 2014 conference on empirical methods in
  natural language processing (EMNLP). pp. 1532--1543 (2014)

\bibitem{peters2018deep}
Peters, M.E., Neumann, M., Iyyer, M., Gardner, M., Clark, C., Lee, K.,
  Zettlemoyer, L.: Deep contextualized word representations. arXiv
  preprint:1802.05365  (2018)

\bibitem{povey2001hugo}
Povey, S., Lovering, R., Bruford, E., Wright, M., Lush, M., Wain, H.: The hugo
  gene nomenclature committee (hgnc). Human genetics  \textbf{109}(6),
  678--680 (2001)

\bibitem{10.1093/bioinformatics/bty845}
Pyysalo, S., Baker, S., Ali, I., Haselwimmer, S., Shah, T., Young, A., Guo, Y.,
  Högberg, J., Stenius, U., Narita, M., Korhonen, A.: {LION LBD: a
  literature-based discovery system for cancer biology}. Bioinformatics
  \textbf{35}(9),  1553--1561 (10 2018)

\bibitem{ray2016rare}
Ray, S., Chiba, N., Yao, C., Guan, X., McConnell, A.M., Brockway, B., Que, L.,
  McQualter, J.L., Stripp, B.R.: Rare sox2+ airway progenitor cells generate
  krt5+ cells that repopulate damaged alveolar parenchyma following influenza
  virus infection. Stem cell reports  \textbf{7}(5),  817--825 (2016)

\bibitem{swanson1990medical}
Swanson, D.R.: Medical literature as a potential source of new knowledge.
  Bulletin of the Medical Library Association  \textbf{78}(1), ~29 (1990)

\bibitem{tshitoyan2019unsupervised}
Tshitoyan, V., Dagdelen, J., Weston, L., Dunn, A., Rong, Z., Kononova, O.,
  Persson, K.A., Ceder, G., Jain, A.: Unsupervised word embeddings capture
  latent knowledge from materials science literature. Nature
  \textbf{571}(7763),  95--98 (2019)

\bibitem{vijay2017virus}
Vijay, R., Fehr, A.R., Janowski, A.M., Athmer, J., Wheeler, D.L., Grunewald,
  M., Sompallae, R., Kurup, S.P., Meyerholz, D.K., Sutterwala, F.S., et~al.:
  Virus-induced inflammasome activation is suppressed by prostaglandin d2/dp1
  signaling. Proceedings of the National Academy of Sciences  \textbf{114}(27),
   E5444--E5453 (2017)

\bibitem{vijay2015critical}
Vijay, R., Hua, X., Meyerholz, D.K., Miki, Y., Yamamoto, K., Gelb, M.,
  Murakami, M., Perlman, S.: Critical role of phospholipase a2 group iid in
  age-related susceptibility to severe acute respiratory syndrome--cov
  infection. Journal of Experimental Medicine  \textbf{212}(11),  1851--1868
  (2015)

\bibitem{zhao2011age}
Zhao, J., Zhao, J., Legge, K., Perlman, S.: Age-related increases in pgd 2
  expression impair respiratory dc migration, resulting in diminished t cell
  responses upon respiratory virus infection in mice. The Journal of clinical
  investigation  \textbf{121}(12),  4921--4930 (2011)

\bibitem{zhou2020pneumonia}
Zhou, P., Yang, X.L., Wang, X.G., Hu, B., Zhang, L., Zhang, W., Si, H.R., Zhu,
  Y., Li, B., Huang, C.L., et~al.: A pneumonia outbreak associated with a new
  coronavirus of probable bat origin. nature  \textbf{579}(7798),  270--273
  (2020)

\bibitem{zhu2020novel}
Zhu, N., Zhang, D., Wang, W., Li, X., Yang, B., Song, J., Zhao, X., Huang, B.,
  Shi, W., Lu, R., et~al.: A novel coronavirus from patients with pneumonia in
  china, 2019. New England Journal of Medicine  (2020)

\end{thebibliography}

\end{document}